\documentclass[11pt,oneside]{llncs}

\usepackage[width=155mm, height=235mm, top=25mm, centering]{geometry}

\usepackage{preprint}
\usepackage{preprintabbrv}

\usepackage{graphicx}
\usepackage{booktabs}
\usepackage{multirow}
\usepackage[accsupp]{axessibility}
\usepackage[pagebackref,breaklinks,colorlinks,citecolor=blue]{hyperref}
\usepackage{orcidlink}
\usepackage{tabularx}
\usepackage{array}

\usepackage{anyfontsize}
\renewcommand{\normalsize}{\fontsize{12}{14}\selectfont}

\newcommand{\layerhead}[2]{L{#1}H{#2}}
\newcommand{\pp}{\,\text{pp}}
\newcommand{\deltaV}{\Delta V}

\definecolor{posgreen}{rgb}{0.0, 0.5, 0.0}
\definecolor{negred}{rgb}{0.7, 0.0, 0.0}

\begin{document}

\title{Locating Demographic Bias at the Attention-Head Level in CLIP's Vision Encoder}

\author{
Alaa Yasser\inst{2} \and
Kittipat Phunjanna\inst{2} \and
Marcos Escudero Vi\~{n}olo\inst{2} \and
Catarina Barata\inst{3} \and
Jenny Benois-Pineau\inst{1}
}

\institute{
University of Bordeaux, France \and
Universidad Autónoma de Madrid, Spain \and
Instituto Superior Técnico, Universidade de Lisboa, Portugal
}

\maketitle

\begin{abstract}
Standard fairness audits of foundation models quantify \textit{that} a model is biased, but not \textit{where} inside the network the bias resides.
We propose a mechanistic fairness audit that combines projected residual-stream decomposition, zero-shot Concept Activation Vectors, and bias-augmented TextSpan analysis to locate demographic bias at the level of individual attention heads in vision transformers. 
As a feasibility case study, we apply this pipeline to the CLIP ViT-L-14 encoder on 42 profession classes of the FACET benchmark, auditing both gender and age bias. For gender, the pipeline identifies four terminal-layer heads whose ablation reduces global bias (Cram\'{e}r's $V$: $0.381 \to 0.362$) while marginally improving accuracy ($+0.42\%$); a layer-matched random control confirms that this effect is specific to the identified heads. 
A single head in the final layer contributes to the majority of the reduction in the most stereotyped classes, and class-level analysis shows that corrected predictions shift toward the correct occupation. 
For age, the same pipeline identifies candidate heads, but ablation produces weaker and less consistent effects, suggesting that age bias is encoded more diffusely than gender bias in this model. 
These results provide preliminary evidence that head-level bias localisation is feasible for discriminative vision encoders and that the degree of localisability may vary across protected attributes.


\end{abstract}
\section{Introduction}
\label{sec:intro}

Foundation models trained on web-scale data have become the backbone for modern multimodal systems~\cite{radford2021learning}. Despite their impressive generalization, these models systematically replicate societal biases embedded in their training corpora~\cite{bender2021dangers, birhane2021multimodal, agarwal2021evaluating}: a CLIP-based occupation classifier misclassifies female doctors as nurses at nearly double the rate of male doctors~\cite{buolamwini2018gender}, yet the standard fairness audit that surfaces this disparity cannot explain \emph{where} it originates inside the network~\cite{mitchell2019model}.

Mechanistic interpretability~\cite{olah2020zoom, elhage2021mathematical} provides tools to look inside the network. The residual-stream view decomposes a transformer's output into additive contributions of individual attention heads, and the TextSpan algorithm~\cite{gandelsman2024interpreting} projects each head's contribution into CLIP's joint text--image space, assigning human-readable labels that describe what the head encodes. These tools have been used to identify heads responsible for colour, texture, or object structure---but never for demographic bias. In the generative setting, Shi~\etal~\cite{Shi_2025_CVPR} showed that bias features in diffusion U-Nets can be located via sparse autoencoders and manipulated for debiasing. However, discriminative vision encoders such as CLIP may route demographic information through different architectural components (attention heads rather than diffusion bottleneck features), and whether bias in these encoders is localisable at the head level has not been addressed.

We bridge this gap by augmenting the TextSpan dictionary with demographic and occupational prototypes, so that bias-related texts compete on equal footing with thousands of general visual descriptions. A head whose variance across the image population is best explained by a gender prototype rather than an occupation or visual-concept text is, by construction, encoding demographic rather than task-relevant information. This bias-augmented TextSpan analysis, combined with zero-shot Concept Projections---a multimodal adaptation of Concept Activation Vectors (CAV)~\cite{kim2018interpretability} that quantifies each head's alignment with demographic versus occupational directions---yields a ranking of the heads most responsible for spurious demographic encoding. We validate these rankings through targeted mean ablation, not as a debiasing strategy but as a causal test: if ablating a head changes predictions in the direction predicted by its demographic ranking, this confirms that the head carries demographic signal that influences classification.

As a feasibility case study, we apply this pipeline to the CLIP ViT-L-14 encoder on the FACET benchmark~\cite{facet2023} (a subset of 42 profession classes), auditing both gender and age bias. For gender, the pipeline identifies a compact set of terminal-layer heads whose ablation reduces global bias while improving accuracy, confirmed against a layer-matched random control. For age, the pipeline identifies candidate heads, but ablation produces weaker and less consistent effects---an outcome that is itself informative, suggesting that age bias is encoded more diffusely than gender bias in this architecture.

Our contribution is threefold:
\begin{enumerate}
    \item[\textbf{(i)}] A diagnostic methodology for locating demographic bias in vision transformers at the attention-head level, combining projected residual-stream decomposition, zero-shot CAV-based head ranking, and bias-augmented TextSpan analysis. The key idea---injecting demographic prototypes into TextSpan's text dictionary so that bias competes with general concepts for variance explained---is, to our knowledge, new.
    \item[\textbf{(ii)}] A feasibility demonstration on CLIP ViT-L-14 showing that the pipeline identifies gender-bias heads whose ablation reduces prediction disparities while improving accuracy, with a layer-matched random control confirming specificity. Class-level analysis shows that corrected predictions shift toward the correct occupation, confirming that the identified heads carry demographic routing signal.
    \item[\textbf{(iii)}] Evidence that the degree of localisability varies across protected attributes: gender bias concentrates in a small set of identifiable heads, while age bias resists the same localisation, pointing to different encoding strategies within the same model.
\end{enumerate}

This work targets the Concept \& Feasibility contribution type. We present the methodology and preliminary evidence that head-level bias localisation is feasible and diagnostically informative, rather than pursuing exhaustive benchmarking or deployment-ready debiasing.

\section{Related Work}
\label{sec:related}
\paragraph{Bias in vision--language models.}
Demographic bias in vision has been studied from face recognition disparities~\cite{buolamwini2018gender} to auditing large-scale vision--language models.
Wang~\etal~\cite{wang2019balanced} and Zhao~\etal~\cite{zhao2017men} showed that even balanced training data propagates bias through learned representations, and Agarwal~\etal~\cite{agarwal2021evaluating} demonstrated that CLIP's zero-shot predictions correlate with racial and gender stereotypes.
Birhane~\etal~\cite{birhane2021multimodal} documented harmful content in large-scale image--text corpora, linking bias partly to the training data.
These studies operate at the \emph{output} level: they measure what the model predicts for different demographic groups, but do not explain which internal components produce the disparity.
The FACET benchmark~\cite{facet2023} provides multi-attribute annotations across occupations and demographic attributes, enabling per-class and per-group fairness analysis.
We use FACET---together with TextSpan---to locate the internal sources of biased predictions.

\paragraph{Mechanistic interpretability.}
The residual-stream hypothesis~\cite{elhage2021mathematical} treats transformers as compositions of additive components that read from and write to a shared representation; the final output can thus be attributed to individual attention heads and MLP blocks.
Gandelsman~\etal~\cite{gandelsman2024interpreting} exploited this additivity in CLIP by projecting each head's write into the joint text--image space via the TextSpan algorithm, producing human-readable labels that describe what each head encodes (e.g., colour, texture, object parts).
In language models, causal tracing~\cite{meng2022locating} and automated circuit discovery~\cite{conmy2023automated} have been used to locate components responsible for factual knowledge, but these techniques have not been applied to fairness in vision encoders.
In the generative domain, Shi~\etal~\cite{Shi_2025_CVPR} proposed DiffLens, which uses sparse autoencoders to disentangle polysemantic neurons in diffusion U-Nets and gradient-based attribution to identify bias-generating features.
Their approach is effective for controlling generation balance, but operates on a different architecture and addresses a different goal (mitigation via feature scaling vs.\ diagnosis of bias localisation).
We work at the attention-head level---an architectural unit that is interpretable through TextSpan without requiring an auxiliary model---and focus on whether bias is localisable in discriminative encoders, rather than assuming localisability and intervening.

\paragraph{Concept-based explanations.}
Kim~\etal~\cite{kim2018interpretability} introduced Concept Activation Vectors (CAVs) as directions in a network's activation space that correspond to user-defined concepts.
A CAV is obtained by training a linear classifier between activations produced by concept examples and random counterexamples; the normal to the decision boundary defines the concept direction.
TCAV then uses directional derivatives along this direction to quantify a concept's influence on predictions.
The approach requires a set of labelled concept images and a probe layer, and operates at the level of an entire layer rather than individual heads.
We adapt the CAV idea to CLIP's multimodal setting in two ways: (a)~concept directions are derived from text embeddings rather than from labelled image sets, making the approach zero-shot, and (b)~we compute concept alignment per head rather than per layer, enabling localisation at a finer architectural granularity.

\paragraph{Bias localisation and debiasing.}
Bolukbasi~\etal~\cite{bolukbasi2016man} proposed subspace projection to debias word embeddings.
Vig~\etal~\cite{vig2020investigating} applied causal mediation analysis to locate gender circuits in GPT-2, showing that specific attention heads mediate gender-biased predictions in language models.
In diffusion models, guidance-based methods~\cite{parihar2024balancing, kwon2023diffusion, li2024self} steer generation toward balanced outputs but do not reveal which internal components drive bias; fine-tuning approaches~\cite{shen2024finetuning} modify model weights globally without component-level attribution.
For discriminative vision encoders, no prior work has attempted to identify individual architectural components responsible for demographic bias.

Our work extends bias localisation to the vision transformer branch of CLIP-like architectures, using zero-shot CAV-based attribution to rank heads by demographic sensitivity and mean ablation to validate them. A key methodological addition is the layer-matched random control: without it, one cannot distinguish targeted bias reduction from the generic effect of removing attention capacity. We note, however, that ablation serves here as a diagnostic instrument, not as a debiasing strategy: our experiments show that neutralising a head that encodes bias toward one demographic value can displace predictions toward another value of the same attribute, leaving the model biased in a different direction rather than unbiased.

\section{Methodology}
\label{sec:method}

We audit the CLIP ViT-L-14 vision encoder~\cite{dosovitskiy2020image}, pretrained on the LAION-2B dataset\footnote{checkpoint \texttt{laion2b\_s32b\_b82k})}~\cite{radford2021learning}, using the projected residual-stream decomposition of Gandelsman~\etal~\cite{gandelsman2024interpreting}. This model contains $L{=}24$ transformer layers, each with $H{=}16$ attention heads, totalling 384 head components.

\subsection{Projected Residual-Stream Decomposition}
The encoder is treated as a residual stream~\cite{elhage2021mathematical}. The final image representation $M_{\text{image}}(I)$ for an input image $I$ is decomposed as:
\begin{equation}
    M_{\text{image}}(I) = P[Z^0]_{\text{cls}} + \sum_{l=0}^{L-1} P[\text{MSA}^l(Z^l)]_{\text{cls}} + \sum_{l=0}^{L-1} P[\text{MLP}^l(\hat{Z}^l)]_{\text{cls}}
\end{equation}
where $P$ denotes the projection matrix mapping from the internal representation space to the joint vision--language embedding space, $Z^0$ is the initial patch embedding, and $\text{MSA}^l$, $\text{MLP}^l$ are the multi-head self-attention and feed-forward blocks at layer $l$, respectively. Each MSA term decomposes into $H$ independent attention heads:
\begin{equation}
    c_{\text{head}}^{l,h} = \sum_{i=0}^{N} P\left(\alpha_i^{l,h} \, W_{VO}^{l,h} \, z_i^l\right)
\end{equation}
where $\alpha_i^{l,h} \in \mathbb{R}$ is the attention weight assigned to token $i$ by head $h$ at layer $l$, $W_{VO}^{l,h}$ is the value-output transition matrix, $z_i^l$ is the residual-stream vector for token $i$ at layer $l$, and $N{=}256$ is the number of spatial patch tokens.

\subsection{Zero-shot CAV-Based Head Ranking}
\label{sec:cav}
We adapt CAV~\cite{kim2018interpretability} to a zero-shot setting by leveraging CLIP's pre-aligned multimodal space. Where CAVs are normally derived from linear probes trained on labelled image sets, we obtain concept directions directly from averaged text embeddings produced by CLIP's text encoder, requiring no additional training data or classifiers.

\paragraph{Text Prototypes.}
Each occupation and demographic attribute is represented by 5 synonym/prototype texts as shown in Table~\ref{tab:prototypes}. Each text is encoded by the CLIP text encoder, $L_2$-normalised, and averaged across the 5 synonyms to produce a stable prototype embedding.

\begin{table}[t]
    \centering
    \caption{Text prototypes used for zero-shot CAV analysis. For occupations, 5~synonyms per class are defined (representative examples shown). For demographics, the full set of prototypes is listed.}
    \label{tab:prototypes}
    \begin{scriptsize}
    \setlength{\tabcolsep}{3pt}
    \renewcommand{\arraystretch}{0.8}
    \begin{tabularx}{\textwidth}{@{}lX@{}}
        \toprule
        \multicolumn{2}{c}{\textbf{Concept Prototypes for Demographic and Occupation Attributes}} \\
        \midrule
        \textbf{Concept} & \textbf{Prototype Texts (5 per concept)} \\
        \midrule
        \multicolumn{2}{@{}l}{\textit{Demographic Attributes (Bias Prototypes)}} \\
        \midrule
        Male      & Male person, Man, Masculine face, Male individual, He \\
        Female    & Female person, Woman, Feminine face, Female individual, She \\
        Non-binary & Non-binary person, Androgynous face, Gender-neutral person, They, Non-binary individual \\
        Young     & Young person, Youth, Young adult, Youthful face, Teenager \\
        Middle    & Middle-aged person, Adult, Mature adult, Middle-aged face, Grown-up \\
        Older     & Older person, Elderly, Senior, Aged face, Elder \\
        \midrule
        \multicolumn{2}{@{}l}{\textit{Occupation Concepts (5 examples of 42)}} \\
        \midrule
        Doctor    & Doctor, Physician, Medical professional, Healthcare provider, MD \\
        Nurse     & Nurse, Healthcare worker, Medical nurse, Caregiver, Registered nurse \\
        Guard     & Guard, Security guard, Watchman, Sentinel, Protector \\
        Dancer    & Dancer, Performer, Ballet dancer, Dance artist, Choreographer \\
        \bottomrule
    \end{tabularx}
    \end{scriptsize}
\end{table}

\paragraph{Occupation--Demographic Alignment Test.}
For each attention head $(l, h)$ and occupation class $p$, we compute a visual centroid $v_{l,h,p}$ by averaging the projected head output $c_{\text{head}}^{l,h}$ across all images of that profession. We then measure how strongly this centroid aligns with the occupation prototype ($v_{\text{occ}}$) versus each demographic prototype ($v_d$) using cosine similarity:
\begin{align}
    S_{\text{occ}}(l, h, p) &= \text{CosSim}(v_{l,h,p},\; v_{\text{occ}}) \label{eq:s_occ}\\
    S_{\text{bias}}(l, h, p, d) &= \text{CosSim}(v_{l,h,p},\; v_{d}) \label{eq:s_bias}
\end{align}
where $v_{\text{occ}}$ and $v_d$ are the averaged CLIP text embeddings for occupation $p$ and the $d$-th demographic dimension (\eg, Male vs.\ Female), respectively. A head whose centroid aligns more strongly with a demographic prototype than with its occupation prototype is a candidate for carrying spurious demographic signal.

\paragraph{Threshold-Based Head Selection.}
A head is flagged as potentially biased for a given profession if its demographic signal is both \emph{directionally specific} and \emph{task-relevant}. Two thresholds are enforced:
\begin{enumerate}
    \item \textbf{Directional specificity} ($\tau_{\text{gap}}$): For head $(l,h)$ and profession $p$, let $d_1$ and $d_2$ be the demographic dimensions with the largest and second-largest absolute bias similarities $|S_{\text{bias}}(l, h, p, d)|$. The directional gap is $G(l, h, p) = |S_{\text{bias}}(l, h, p, d_1)| - |S_{\text{bias}}(l, h, p, d_2)|$. We require $G(l, h, p) > \tau_{\text{gap}}$. This retains heads that respond to a specific demographic direction rather than firing uniformly for all prototypes.
    \item \textbf{Task relevance} ($\tau_{\text{occ}}$): We require $|S_{\text{occ}}(l, h, p)| > \tau_{\text{occ}}$, filtering out heads with negligible occupation alignment, whose relative bias scores would otherwise be unstable.
\end{enumerate}
Each head $(l, h)$ that passes both thresholds for at least one profession is treated as a candidate bias head.

\paragraph{Grid Search for Threshold Selection.}
We sweep  $\tau_{\text{gap}}$ and $\tau_{\text{occ}}$ over a $40 \times 60$ grid ($\tau_{\text{gap}} \in [0.005, 0.20]$, $\tau_{\text{occ}} \in [0.005, 0.30]$, step $0.005$). For each combination, all heads passing both thresholds are mean-ablated (Sec.~\ref{sec:ablation}), and we record (i) the average Cram\'{e}r's $V$ across statistically significant classes and (ii) overall accuracy across all 42 classes. The selected threshold pair is the one that minimises $V$ subject to accuracy not declining.

This selection procedure uses the evaluation metric, which introduces a risk of circularity. Three factors mitigate this concern: (a)~the grid search selects \emph{threshold values}, not individual heads---heads are determined by the CAV alignment scores, which are computed independently of the ablation outcome; (b)~the layer-matched random control (Sec.~\ref{sec:ablation}) provides an independent check, since if the selected heads were arbitrary, random heads from the same layers would produce comparable $\Delta V$; and (c)~the bias-augmented TextSpan analysis (Sec.~\ref{sec:textspan}) and the class-level prediction redistribution (Sec.~\ref{sec:class_level}) provide corroboration that is independent of the grid search objective.

\subsection{Bias-Augmented TextSpan Analysis}
\label{sec:textspan}

To generate human-readable semantic annotations for each attention head, we apply the TextSpan algorithm~\cite{gandelsman2024interpreting} using an augmented dictionary. We extend TextSpan's original 3,497 general visual concepts with 42 occupation and 6 demographic embeddings (averaged across synonyms/prototypes and 80 ImageNet templates~\cite{radford2021learning}) , totaling 3,545 texts. For each head, projected attention features $c_{\text{head}}^{l,h}$ are extracted across all FACET images and TextSpan's rank-80 SVD iterative-removal process is used to identify the top-$K$ ($K{=}20$) explanatory texts that maximize cross-image variance. If a demographic prototype (\eg, \texttt{gender\_female}) surfaces among these top texts, it provides independent qualitative corroboration of the quantitative CAV ranking, offering converging evidence of demographic encoding.

\subsection{Statistical Bias Quantification}
\label{sec:stat_tests}

We quantify bias in the model's predictions (before and after ablation) using the prediction contingency table $O \in \mathbb{R}^{G \times K}$, where $G$ is the number of demographic groups and $K$ the number of predicted classes.

\paragraph{Minimum Group Size.}
All statistical tests require a minimum of 20 images per demographic group within each class. For gender, the Non-Binary group falls below this threshold in every profession class and is therefore excluded from the chi-squared analyses. Across all 42 classes the Non-Binary count never reaches 20: the largest class (singer) contains only 10 Non-Binary images, and the entire evaluation set comprises only 55 Non-Binary images in total. This is a limitation of the FACET dataset's demographic distribution; the CAV pipeline (Sec.~\ref{sec:cav}) does use Non-Binary prototypes in the head-identification step.

\paragraph{Chi-Squared Test and Cram\'{e}r's $V$.}
For each true profession class, we test $H_0$: the distribution over predicted classes is identical across demographic groups. The chi-squared statistic is:
\begin{equation}
    \chi^2 = \sum_{g=1}^{G}\sum_{k=1}^{K} \frac{(O_{gk} - E_{gk})^2}{E_{gk}}
\end{equation}
where $O_{gk}$ is the observed count and $E_{gk} = (R_g \cdot C_k)/N$ is the expected count under independence ($R_g$: row total, $C_k$: column total, $N$: grand total). We apply the Benjamini--Hochberg (BH) correction~\cite{benjamini1995controlling} across all 42 class-level tests to control the false discovery rate at $\alpha{=}0.05$. As effect-size measure independent of sample size, we report Cram\'{e}r's $V$~\cite{cramer1946contribution}:
\begin{equation}
    V = \sqrt{\frac{\chi^2}{N \cdot (\min(G,K) - 1)}}
\end{equation}

\subsection{Mean Ablation}
\label{sec:ablation}
To validate identified heads, we perform mean ablation~\cite{vig2020investigating}: the projected output of a targeted head $c_{\text{head}}^{l,h}$ is replaced by its mean $\bar{c}_{\text{head}}^{l,h}$ computed across all images in the evaluation set. This neutralises the head's input-specific contribution while preserving its average effect on the residual stream. Mean ablation serves here as a diagnostic instrument: if removing a head's input-specific signal changes predictions in the direction predicted by its demographic ranking, this confirms that the head carries demographic information that influences classification. It is not proposed as a debiasing strategy.

\paragraph{Layer-Matched Random Control.}
To establish that any observed effect is specific to the identified heads, we compare against layer-matched random heads: the same number of heads drawn from the same layer distribution as the suspected set (\eg, L21:~2, L22:~1, L23:~1). The experiment is repeated 10 times with different random seeds; means and standard deviations are reported.

\section{Experimental Setup}
\label{sec:setup}

\subsection{Dataset: FACET Benchmark}

We evaluate on the FACET benchmark~\cite{facet2023}, removing 10 temporary prediction-sink classes (Patient, Backpacker, Computer user, Student, Prayer, Climber, Runner, Skateboarder, Cheerleader, Speaker) from both the image pool and prediction targets. This exclusion yields a \textbf{42-class} evaluation set of \textbf{25,416 images}, each annotated for Gender (Male, Female, Non-Binary) and Age (Young, Middle, Older).

\subsection{Feature Extraction and Classification}
Per-head attention features and MLP features are extracted from the ViT-L-14 encoder for each image. The classification is zero-shot: we project the sum of all heads' contributions and MLP outputs through a TextSpan-derived classifier~\cite{gandelsman2024interpreting} to produce logits over the 42 profession classes. Baseline accuracy is \textbf{64.30\%}.

\section{Global Analysis}
\label{sec:global}

\subsection{Baseline Bias Prevalence}
\label{sec:bias_detection}

Baseline evaluation reveals pervasive gender bias but circumscribed age bias. Of 42 profession classes, \textbf{19} exhibit statistically significant gender disparities (BH-corrected $p < 0.05$), whereas only \textbf{7} show significant age bias, foreshadowing an asymmetry observed throughout our experiments. Table~\ref{tab:bias_detail_combined} details the most affected classes. The stereotyped doctor-to-nurse confusion is highly emblematic: 78.2\% of female doctors are misclassified as nurses (vs.\ 39.4\% for males), while only 13.4\% are correctly classified as doctors (vs.\ 37.2\% for males).

\begin{table*}[t]
    \centering
    \caption{Top-5 gender- and age-biased occupation classes. The left half shows Gender Bias and the right half shows Age Bias ranked by Cram\'{e}r's $V$, including top-3 predicted classes with per-group rates (\%). The correct class is highlighted.}
    \label{tab:bias_detail_combined}
    \begin{scriptsize}
    \setlength{\tabcolsep}{3.2pt}
    \renewcommand{\arraystretch}{1.1}

    \begin{tabular}{@{} l c l cc c l c l cc @{}}
        \toprule
        \multicolumn{5}{c}{\textbf{GENDER BIAS}} & & \multicolumn{5}{c}{\textbf{AGE BIAS}} \\
        \cmidrule(lr){1-5} \cmidrule(lr){7-11}
        \textbf{Class} & \textbf{$V$} & \textbf{$\to$Pred.} & \textbf{Male} & \textbf{Fem.} & & \textbf{Class} & \textbf{$V$} & \textbf{$\to$Pred.} & \textbf{Yng.} & \textbf{Old.} \\
        \midrule
        
        \multirow[t]{3}{*}{Nurse} & \multirow[t]{3}{*}{0.450} 
            & \textbf{Nurse}     & 44.9 & 83.1 & & \multirow[t]{3}{*}{Guard} & \multirow[t]{3}{*}{0.238} 
            & \textbf{Guard}     & 89.3 & 44.8 \\
            & & Doctor           & 20.1 & 3.1  & & & & Laborer           & 0.6  & 10.3 \\
            & & Repairman        & 7.7  & 0.4  & & & & Seller            & 1.3  & 6.9 \\
        \midrule
        
        \multirow[t]{3}{*}{Doctor} & \multirow[t]{3}{*}{0.399} 
            & Nurse             & 39.4 & 78.2 & & \multirow[t]{3}{*}{Ballplayer} & \multirow[t]{3}{*}{0.186} 
            & \textbf{Ballplayer}& 68.2 & 83.3 \\
            & & \textbf{Doctor}  & 37.2 & 13.4 & & & & Basket. pl.       & 10.3 & 0.0 \\
            & & Seller           & 5.0  & 2.8  & & & & Tennis player     & 7.0  & 0.0 \\
        \midrule
        
        \multirow[t]{3}{*}{Laborer} & \multirow[t]{3}{*}{0.298} 
            & \textbf{Laborer}  & 44.7 & 29.8 & & \multirow[t]{3}{*}{Nurse} & \multirow[t]{3}{*}{0.182} 
            & \textbf{Nurse}     & 80.4 & 48.6 \\
            & & Farmer           & 9.6  & 22.1 & & & & Doctor            & 4.3  & 28.6 \\
            & & Seller           & 3.8  & 13.0 & & & & Lifeguard         & 3.3  & 6.0 \\
        \midrule
        
        \multirow[t]{3}{*}{Craftsman} & \multirow[t]{3}{*}{0.279} 
            & \textbf{Craftsman}& 41.5 & 31.6 & & \multirow[t]{3}{*}{Seller} & \multirow[t]{3}{*}{0.129} 
            & \textbf{Seller}    & 66.9 & 82.8 \\
            & & Seller           & 14.2 & 38.6 & & & & Retailer          & 23.2 & 2.6 \\
            & & Painter          & 11.4 & 14.6 & & & & Painter           & 0.0  & 5.2 \\
        \midrule
        
        \multirow[t]{3}{*}{Ballplayer} & \multirow[t]{3}{*}{0.255} 
            & \textbf{Ballplayer}& 82.4 & 33.3 & & \multirow[t]{3}{*}{Boatman} & \multirow[t]{3}{*}{0.126} 
            & \textbf{Boatman}   & 70.5 & 90.7 \\
            & & Tennis player    & 3.4  & 30.0 & & & & Lifeguard         & 28.5 & 7.2 \\
            & & Referee          & 5.3  & 13.3 & & & & Farmer            & 0.4  & 1.0 \\
        \bottomrule
    \end{tabular}
    \end{scriptsize}
\end{table*}

\subsection{Identified Bias Heads}
\label{sec:head_ranking}

\subsection{Global Ablation with Random Control}
\label{sec:global_ablation}

The result for Layer-Matched ablation is shown in Table~\ref{tab:global_combined}. The suspected ablation improves accuracy for both cases while being higher than random heads simultaneously. The gender $V$ shows huge drop by 5\% relative while age $V$ shows extremely small improvement.

\begin{table}[t]
    \centering
    \caption{Global ablation results for gender-head and age-head experiments, each vs.\ a layer-matched random control ($\pm$ std over 10 runs). $V$ refers to the target attribute of each experiment.}
    \label{tab:global_combined}
    \begin{scriptsize}
    \setlength{\tabcolsep}{3pt}
    \renewcommand{\arraystretch}{0.8}
    \begin{tabular}{@{}lccc@{}}
        \toprule
        \multicolumn{4}{c}{\textbf{Global Bias Reduction ($\deltaV$): Suspected vs.\ Random Control}} \\
        \midrule
        \textbf{Condition} & \textbf{Accuracy} & \textbf{Cram\'{e}r's $V$} & \textbf{$\deltaV$} \\
        \midrule
        \multicolumn{4}{@{}l}{\textit{Gender-Head Experiment ($V$ measured on gender)}} \\
        \midrule
        Baseline             & 64.30\%                        & 0.381                        & ---                 \\
        Suspected (4)        & 64.72\%                        & 0.362                        & \textcolor{posgreen}{$-0.019$} \\
        Random avg (4)       & 64.16\%\,$\pm$\,0.27           & 0.381\,$\pm$\,0.002          & \textcolor{negred}{$-0.000$} \\
        \midrule
        \multicolumn{4}{@{}l}{\textit{Age-Head Experiment ($V$ measured on age)}} \\
        \midrule
        Baseline             & 64.30\%                        & 0.224                        & ---                 \\
        Suspected (3)        & 64.50\%                        & 0.222                        & \textcolor{posgreen}{$-0.002$} \\
        Random avg (3)       & 63.87\%\,$\pm$\,0.45           & 0.226\,$\pm$\,0.001          & \textcolor{negred}{$+0.002$} \\
        \bottomrule
    \end{tabular}
    \end{scriptsize}
\end{table}

\section{Class-Level Analysis}
\label{sec:class_level}

\begin{figure*}[t]
    \centering
    \includegraphics[width=\textwidth]{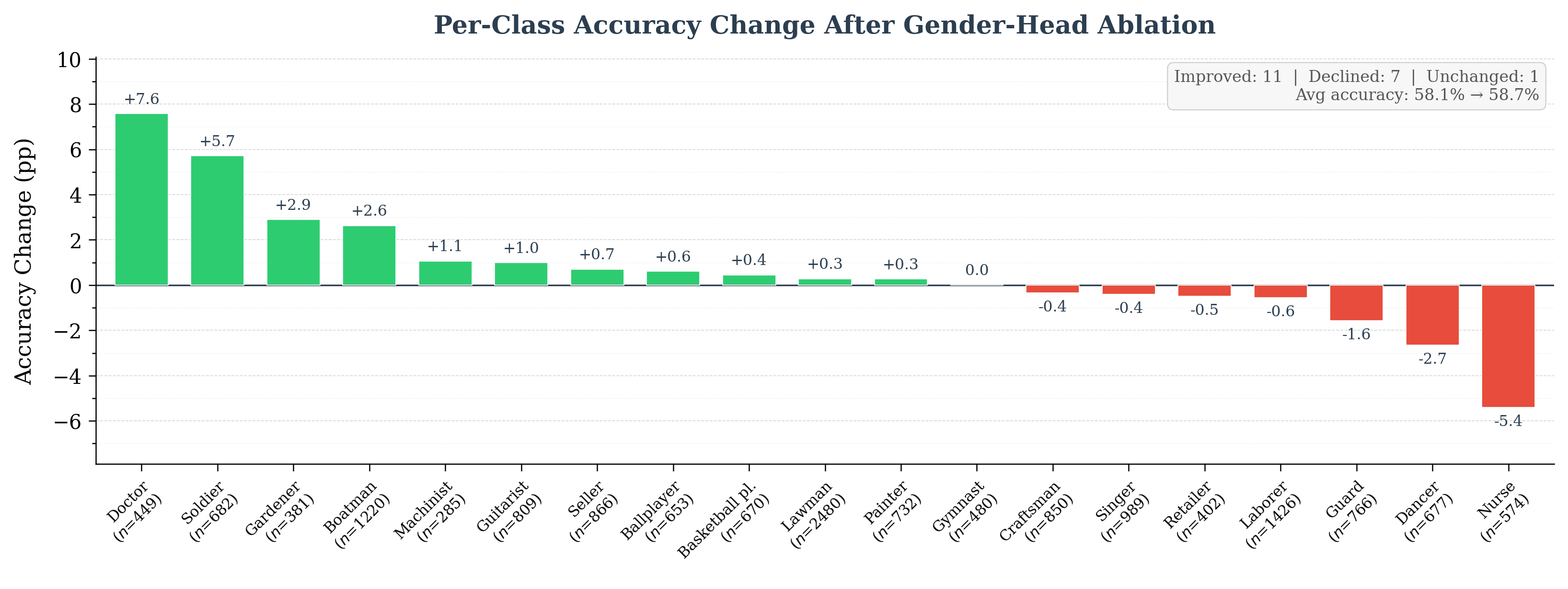}
    \caption{Per-class accuracy change (in percentage points) after combined gender-head ablation across all 19 classes with statistically significant gender bias (BH-corrected $p < 0.05$). Green bars indicate improved accuracy; red bars indicate declines. Sample sizes ($n$) are shown below each class label.}
    \label{fig:perclass_gender}
\end{figure*}

While global metrics confirm the aggregate effect, class-level analysis reveals where bias reduction concentrates and which individual heads are responsible. Figure~\ref{fig:perclass_gender} shows that ablation improves accuracy for 11 classes, with the largest gains in doctor (+7.6\,pp) and soldier (+5.7\,pp). The largest drop occurs in nurse ($-5.4$\,pp), which is the expected trade-off: the same heads that routed female doctors to nurse also inflated nurse accuracy for male images. Average accuracy across all 19 significant classes is preserved (58.1\% $\to$ 58.7\%), confirming that gender-head ablation redistributes predictions toward fairer outcomes without degrading overall performance. We then examine three exemplar classes in detail.

\subsection{Doctor Class (Gender)}
\label{sec:class_doctor}

As noted in Sec.~\ref{sec:bias_detection}, the doctor$\to$nurse confusion is the most emblematic stereotyped prediction, with a baseline gap of 38.8\pp. Table~\ref{tab:doctor_combined} shows the effect of ablating each suspected gender head individually and combined, with both bias metrics and per-gender prediction redistribution. \layerhead{23}{4} alone accounts for $\deltaV = -0.162$ out of the combined $-0.187$ (\textbf{87\%} of the total bias reduction) and lifts female doctor accuracy from 13.4\% to 26.3\%. Rescued predictions go primarily to the correct class (doctor), confirming that the ablation suppresses genuine gender-driven misrouting rather than introducing random noise. However, this also means, conversely, nurse images are predicted as doctor more.

\begin{table*}[t]
    \centering
    \caption{Doctor class: per-head and combined ablation results (gender). Left columns show bias metrics; right columns show prediction redistribution by gender. Heads sorted by $\Delta V$.}
    \label{tab:doctor_combined}
    \begin{scriptsize}
    \setlength{\tabcolsep}{4pt} 
    \renewcommand{\arraystretch}{1.1} 
    
    \newcolumntype{C}[1]{>{\centering\arraybackslash}p{#1}}

    \begin{tabular}{@{} l cc C{5.0em} C{5.0em} C{5.0em} C{5.0em} @{}}
        \toprule
        \multicolumn{7}{c}{\textbf{Doctor Class --- Per-Head Ablation and Prediction Redistribution (Gender)}} \\
        \midrule
        & & & \multicolumn{2}{c}{\textbf{Male} ($N{=}180$)} & \multicolumn{2}{c}{\textbf{Female} ($N{=}179$)} \\
        \cmidrule(lr){4-5} \cmidrule(lr){6-7}
        \textbf{Condition} & \textbf{$V$} & \textbf{$\Delta V$} & \textbf{$\to$Doctor} & \textbf{$\to$Nurse} & \textbf{$\to$Doctor} & \textbf{$\to$Nurse} \\
        \midrule
        Baseline              & 0.399 & ---      & 37.2 & 39.4 & 13.4 & 78.2 \\
        \midrule
        \layerhead{23}{4}     & 0.237 & $-0.162$ & 38.3 & 37.8 & \textbf{26.3} & 60.9 \\
        \layerhead{21}{2}     & 0.355 & $-0.044$ & 36.7 & 39.4 & 14.0 & 74.3 \\
        \layerhead{21}{10}    & 0.356 & $-0.043$ & 37.8 & 38.9 & 16.2 & 73.7 \\
        \layerhead{22}{14}    & 0.402 & $+0.003$ & 36.7 & 39.4 & 13.4 & 78.8 \\
        \midrule
        \textbf{Combined (4)} & 0.212 & $-0.187$ & 38.3 & 36.7 & \textbf{29.6} & \textbf{55.3} \\
        \bottomrule
    \end{tabular}
    \end{scriptsize}
\end{table*}

\subsection{Craftsman Class (Gender)}
\label{sec:class_craftsman}

Female craftsman images are predicted as \textit{seller} at 38.6\% versus 14.2\% for males ($\Delta = 24.4$\pp). Table~\ref{tab:craftsman_combined} shows the effect of ablating each suspected gender head individually and combined, with both bias metrics and per-gender prediction redistribution. \layerhead{23}{4} again dominates, accounting for $\Delta V = -0.053$ out of the combined $-0.059$ (\textbf{90\%} of the total bias reduction), demonstrating that L23H4's gender-routing role generalizes beyond the doctor--nurse axis.

\begin{table*}[t]
    \centering
    \caption{Craftsman class: per-head and combined ablation results (gender). Left columns show bias metrics; right columns show prediction redistribution by gender. Heads sorted by $\Delta V$.}
    \label{tab:craftsman_combined}
    \begin{scriptsize}
    \setlength{\tabcolsep}{4pt}
    \renewcommand{\arraystretch}{1.1}

    \newcolumntype{C}[1]{>{\centering\arraybackslash}p{#1}}

    \begin{tabular}{@{} l cc C{5.0em} C{5.0em} C{5.0em} C{5.0em} @{}}
        \toprule
        \multicolumn{7}{c}{\textbf{Craftsman Class --- Per-Head Ablation and Prediction Redistribution (Gender)}} \\
        \midrule
        & & & \multicolumn{2}{c}{\textbf{Male} ($N{=}598$)} & \multicolumn{2}{c}{\textbf{Female} ($N{=}158$)} \\
        \cmidrule(lr){4-5} \cmidrule(lr){6-7}
        \textbf{Condition} & \textbf{$V$} & \textbf{$\Delta V$} & \textbf{$\to$Craftsman} & \textbf{$\to$Seller} & \textbf{$\to$Craftsman} & \textbf{$\to$Seller} \\
        \midrule
        Baseline              & 0.279 & ---      & 41.5 & 14.2 & 31.6 & 38.6 \\
        \midrule
        \layerhead{23}{4}     & 0.226 & $-0.053$ & 38.6 & 15.4 & \textbf{39.2} & 32.3 \\
        \layerhead{21}{2}     & 0.273 & $-0.007$ & 41.3 & 14.2 & 32.3 & 38.0 \\
        \layerhead{21}{10}    & 0.273 & $-0.007$ & 41.6 & 14.4 & 31.0 & 38.6 \\
        \layerhead{22}{14}    & 0.275 & $-0.004$ & 41.6 & 14.2 & 30.4 & 38.0 \\
        \midrule
        \textbf{Combined (4)} & 0.220 & $-0.059$ & 38.6 & 15.2 & \textbf{41.1} & \textbf{31.0} \\
        \bottomrule
    \end{tabular}
    \end{scriptsize}
\end{table*}

\begin{figure}[t]
    \centering
    \begin{tabular}{cccc}
    \includegraphics[width=0.25\textwidth]{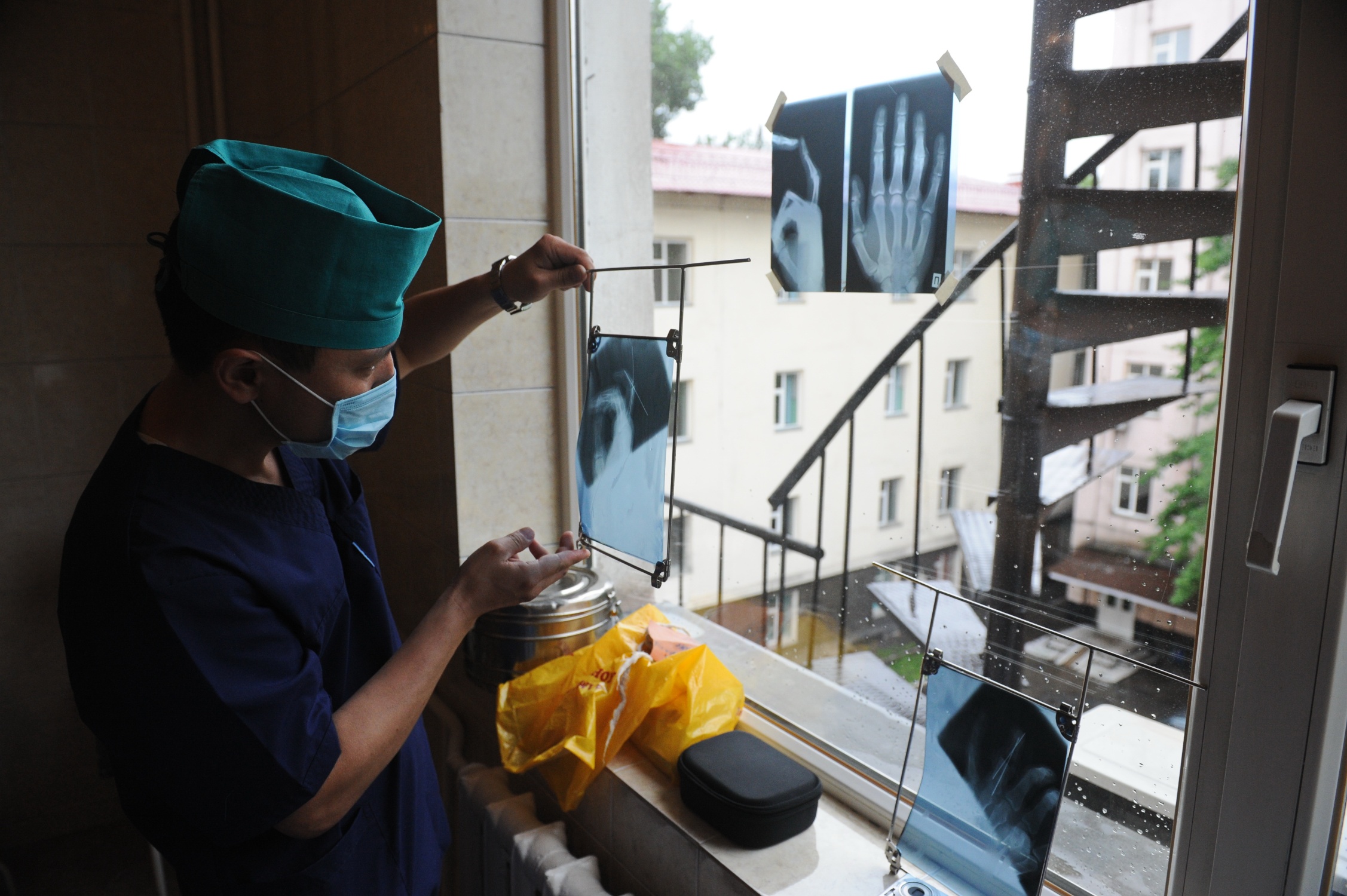} &
    \includegraphics[width=0.25\textwidth]{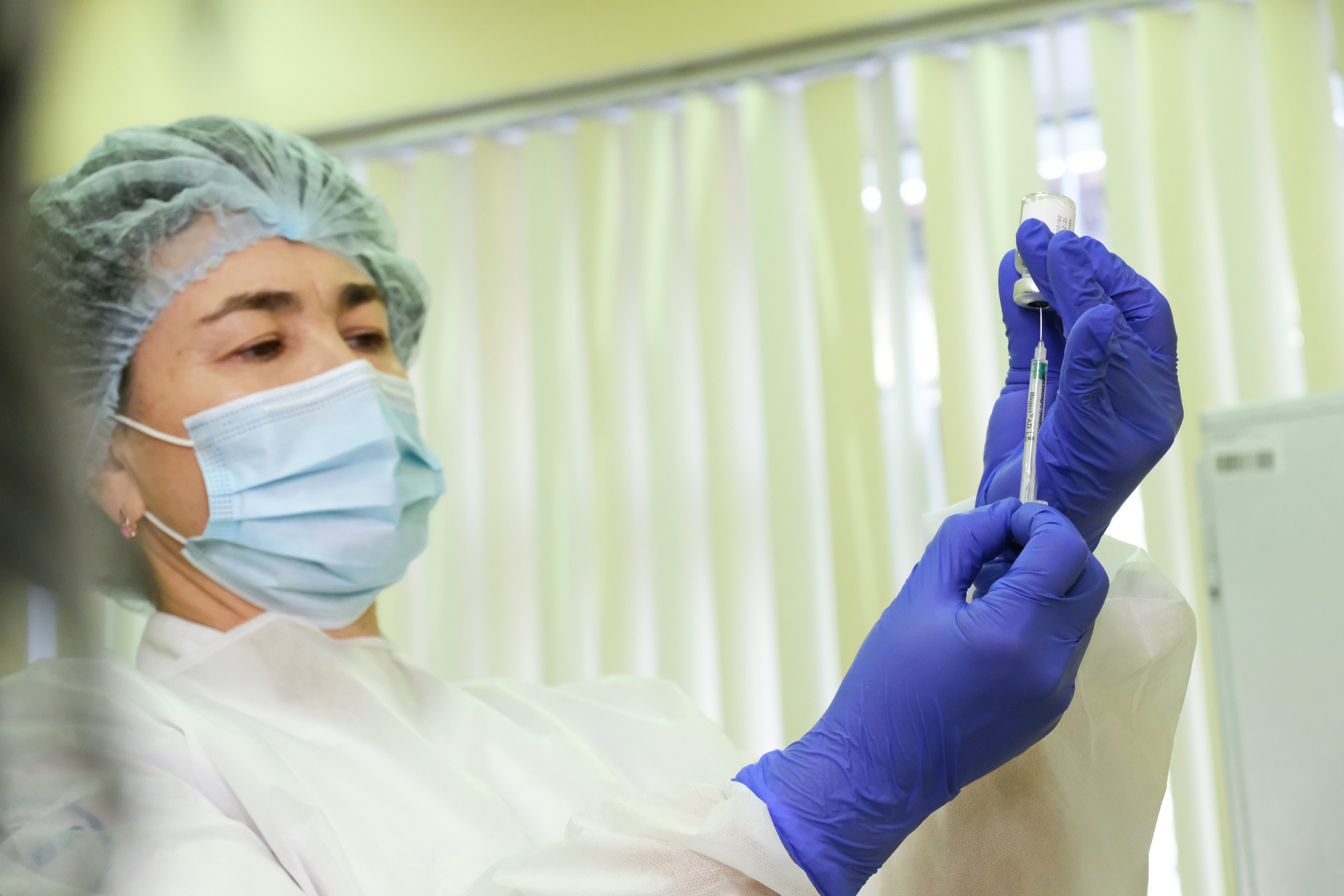} &
    \includegraphics[width=0.25\textwidth]{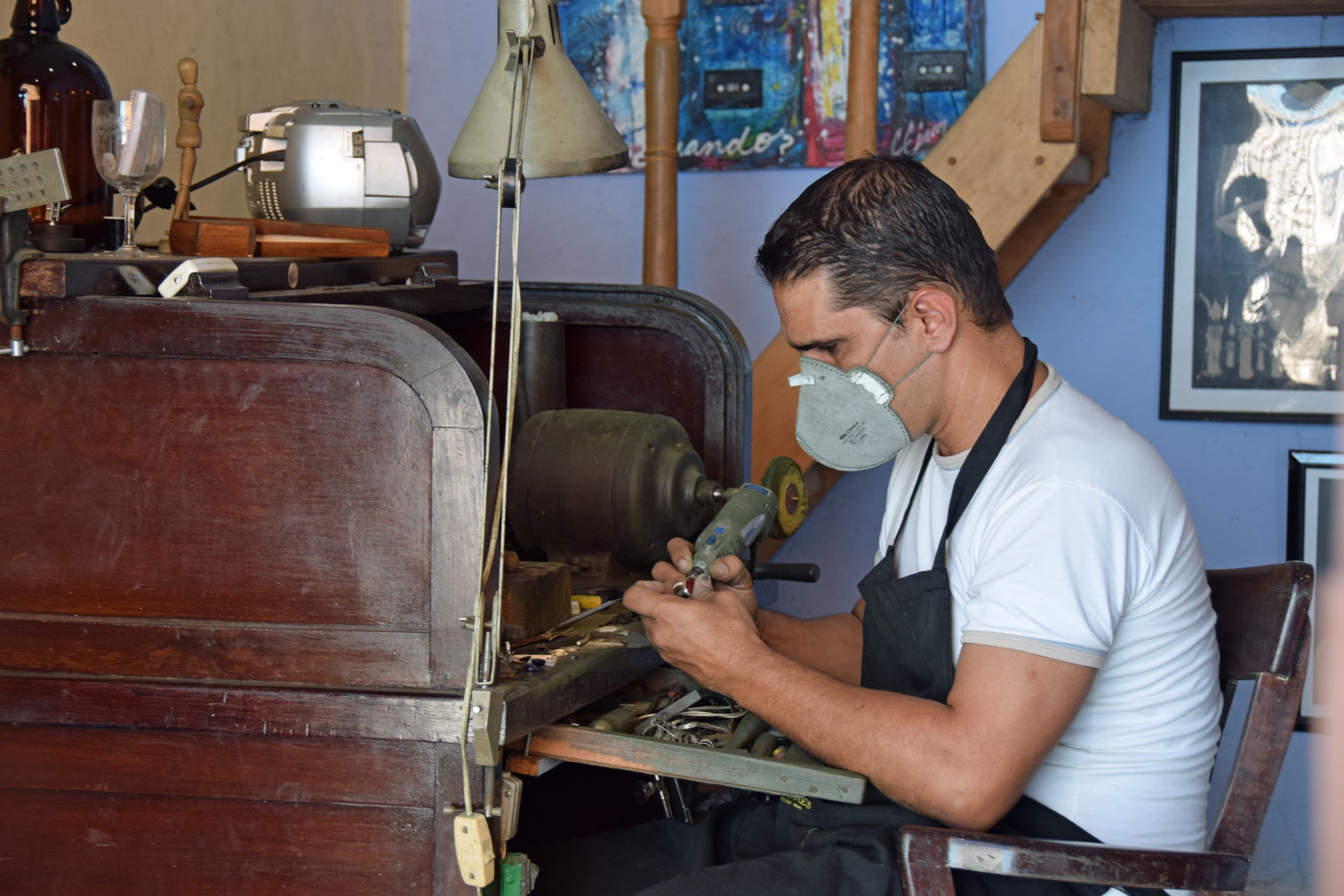} &
    \includegraphics[width=0.25\textwidth]{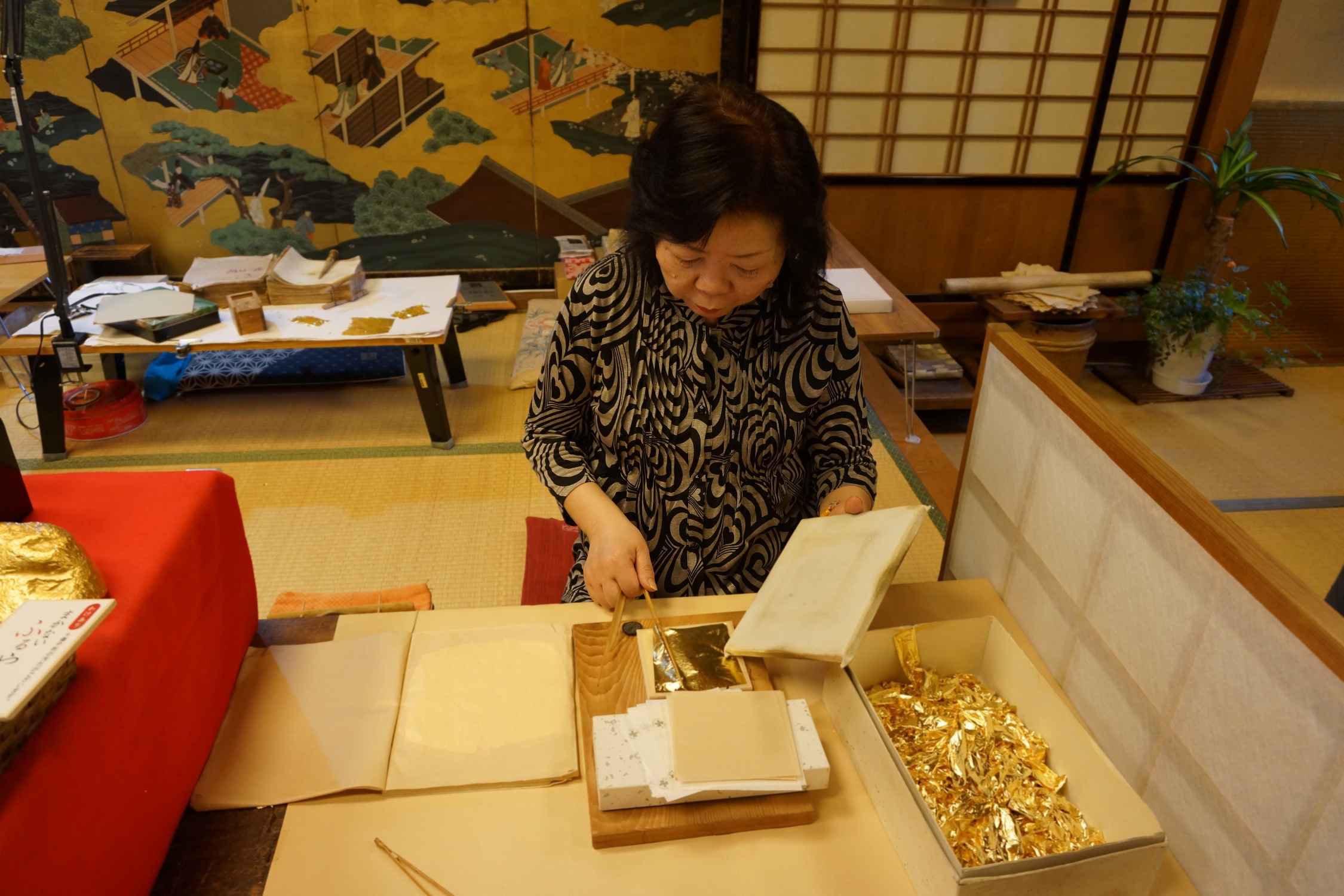} \\
    (a) Doctor (Male) & (b) Doctor (Female) & (c) Craftsman (Male) & (d) Craftsman (Female)
\end{tabular}
    \caption{Qualitative examples of correction after gender-bias head ablation. All images were misclassified at baseline (e.g., nurse for doctor, machinist or seller for craftsman) and correctly reclassified after ablating the top-10 identified gender heads. The images are chosen from cases where the demographic-parity gap was most pronounced.}
    \label{fig:corrections}
\end{figure}

\paragraph{Qualitative Impact of Ablation.}
To understand how the ablation of identified heads affects model behavior image-by-image, we inspect samples that were misclassified at baseline but corrected after ablation. Figure~\ref{fig:corrections} shows examples from the \textit{doctor} and \textit{craftsman} classes. In these cases, the baseline model appears to have over-relied on gendered visual cues, predicting nurse for female doctors or machinist/seller for craftsmen. By neutralizing the terminal-layer bias heads, the model's prediction redirects toward the correct occupational class, demonstrating a functional shift in the decision routing for these specific instances.

\subsection{Guard Class (Age)}
\label{sec:class_guard}

Table~\ref{tab:guard_combined} evaluates the \textit{guard} class, which exhibits the most pronounced baseline age bias (89.3\% accuracy for young vs.\ 44.8\% for older individuals). In contrast to the gender ablation results, neutralizing the suspected age heads fails to mitigate this disparity. Individually, only \layerhead{21}{5} yields a negligible reduction ($\Delta V = -0.004$) , while \layerhead{22}{4} and \layerhead{23}{4} exacerbate the bias. Consequently, combined ablation slightly increases overall bias ($\Delta V = +0.009$) without improving older guard accuracy. This divergence confirms that age bias encoding is highly diffuse; unlike the concentrated gender-routing mechanisms (e.g., \layerhead{23}{4}), ablating the top CAV-selected age heads is insufficient for targeted debiasing.

\begin{table*}[t]
    \centering
    \caption{Guard class: per-head and combined ablation results (age). Left columns show bias metrics; right columns show top-2 predictions by age group (top confusion per group: Gardener for Young, Reporter for Middle, Seller for Older). Heads sorted by $\Delta V$.}
    \label{tab:guard_combined}
    \begin{scriptsize}
    \setlength{\tabcolsep}{2.5pt} 
    \renewcommand{\arraystretch}{1.2}
    \newcolumntype{Y}{>{\centering\arraybackslash}X}
    \begin{tabularx}{\textwidth}{@{} l cc YYYYYY @{}}
        \toprule
        \multicolumn{9}{c}{\textbf{Guard Class --- Per-Head Ablation and Prediction Redistribution (Age)}} \\
        \midrule
        & & & \multicolumn{2}{c}{\textbf{Young} ($N{=}159$)} & \multicolumn{2}{c}{\textbf{Middle} ($N{=}460$)} & \multicolumn{2}{c}{\textbf{Older} ($N{=}29$)} \\
        \cmidrule(lr){4-5} \cmidrule(lr){6-7} \cmidrule(lr){8-9}
        \textbf{Condition} & \textbf{$V$} & \textbf{$\Delta V$} & \textbf{$\to$Guard} & \textbf{$\to$Garden.} & \textbf{$\to$Guard} & \textbf{$\to$Report.} & \textbf{$\to$Guard} & \textbf{$\to$Seller} \\
        \midrule
        Baseline              & 0.238 & ---      & 89.3 & 1.3 & 69.1 & 3.5 & 44.8 & 6.9 \\
        \midrule
        \layerhead{21}{5}     & 0.234 & $-0.004$ & 88.7 & 1.3 & 68.7 & 3.9 & 44.8 & 10.3 \\
        \layerhead{22}{4}     & 0.246 & $+0.008$ & 90.6 & 1.3 & 70.0 & 3.0 & 44.8 & 10.3 \\
        \layerhead{23}{4}     & 0.252 & $+0.015$ & 89.3 & 1.3 & 67.6 & 3.9 & 41.4 & 6.9 \\
        \midrule
        \textbf{Combined (3)} & 0.247 & $+0.009$ & 89.9 & 1.3 & 68.3 & 3.9 & 44.8 & 10.3 \\
        \bottomrule
    \end{tabularx}
    \end{scriptsize}
\end{table*}

\section{Discussion}
\label{sec:discussion}

\paragraph{Gender bias concentrates in a small set of terminal heads.}
Ablating four identified heads (1.0\% of all attention heads) reduces global gender bias ($V$: $0.381 \to 0.362$, $\Delta V = -0.019$) while slightly improving overall accuracy ($64.30\% \to 64.72\%$). A layer-matched random ablation of the same size produces no comparable change ($\Delta V = -0.000 \pm 0.002$), confirming that the effect is specific to the identified heads rather than a generic reduction in attention capacity.

\paragraph{The effect is dominated by a single head.}
Within the identified set, L23H4 accounts for $\Delta V = -0.162$ of the combined $-0.187$ in the doctor class (87\%) and raises female doctor accuracy from 13.4\% to 26.3\%. A similar concentration appears in the craftsman class, where L23H4 explains roughly 90\% of the combined bias reduction. The remaining three heads contribute smaller additive effects. This concentration in a single final-layer head suggests that gender-related prediction routing passes through a narrow bottleneck just before the classification output.

\paragraph{Prediction redistribution and accuracy trade-offs.}
Ablation does not uniformly improve accuracy. Of the 19 significantly gender-biased classes, 11 improve while 7 decline and 1 is unchanged. The largest decrease occurs in nurse ($-5.4$\,pp), which is the mirror of the improvement in doctor: at baseline, the model routes female doctor images to nurse, inflating nurse accuracy. After ablation this routing weakens, improving doctor predictions at the expense of nurse. This trade-off illustrates that the identified heads carry demographic signal that the model uses for classification---removing that signal redistributes predictions rather than creating a neutral classifier. Mean ablation is therefore informative as a diagnostic tool but insufficient as a debiasing strategy: neutralising a head that encodes bias toward one demographic value can displace predictions toward another value of the same attribute, leaving the model biased in a different direction rather than unbiased.

\paragraph{Age bias is not localised by head-level analysis.}
In contrast to gender, the three CAV-identified age heads produce only a small global change ($\Delta V = -0.003$) and do not reduce the largest age disparity observed in the guard class. Individual ablations yield negligible or inconsistent effects, and combined ablation slightly increases bias in that class ($\Delta V = +0.009$). These results suggest that age-related prediction behaviour is not concentrated in a small number of heads, at least under the localisation procedure used here. Whether this reflects a genuinely diffuse encoding or a limitation of the zero-shot CAV ranking for age attributes remains an open question.

\paragraph{Cross-attribute entanglement.}
L23H4 appears in both the gender and age rankings and receives TextSpan annotations related to gender descriptors (\textit{A photo of a woman}, \textit{Gender male}). Its ablation reduces gender bias while slightly increasing age bias in the guard class, suggesting that this head encodes demographic information that cuts across attribute boundaries rather than being tied to a single protected dimension. The age heads' cross-effect on gender ($\Delta V_{\text{gender}} = -0.015$ when ablating age heads) is largely driven by L23H4's presence in both sets.

\paragraph{Limitations.}
Our analysis focuses on the terminal layers (20--23) of ViT-L-14, where mean ablation produces measurable effects~\cite{gandelsman2024interpreting}. Earlier layers are already close to their mean behaviour across images, so heads that pass demographic information forward without large input-specific variance will not be detected. The head ranking is derived from the FACET benchmark and may miss heads encoding demographic signals for occupations not represented in the dataset. The grid search for threshold selection uses the evaluation metric (Cram\'{e}r's $V$), introducing a risk of circularity mitigated by the independent random control, TextSpan corroboration, and class-level redistribution analysis (see Sec.~\ref{sec:method}). The Non-Binary group ($N{=}55$) is too sparse for per-class statistical testing, restricting validation to the Male--Female comparison. Finally, mean ablation removes the entire head contribution rather than selectively suppressing its demographic component, limiting the granularity of the analysis.

\section{Conclusion}
\label{sec:conclusion}

We presented a diagnostic methodology to locate demographic bias at the attention-head level in the CLIP ViT-L-14 vision encoder, combining projected residual-stream decomposition, zero-shot CAV-based head ranking, and bias-augmented TextSpan analysis. Applied to a subset of 42 profession classes of the FACET benchmark, the pipeline identifies four gender-bias heads in the terminal layers whose ablation reduces global gender bias while improving accuracy, confirmed against a layer-matched random control. A single head in the final layer (L23H4) dominates the effect, and class-level analysis shows that corrected predictions shift toward the correct occupation. For age, the same pipeline identifies candidate heads, but ablation produces weaker and less consistent effects, suggesting that age bias is encoded more diffusely in this architecture.

Two findings deserve emphasis. First, ablation confirms that the identified heads carry demographic signal used by the classifier, but it does not produce a neutral model: reducing bias for one class (doctor) displaces it to another (nurse). This underscores that mean ablation is a diagnostic instrument, not a debiasing strategy, and that future work on intervention must account for redistribution effects. Second, the contrast between gender and age localisation suggests that different protected attributes may require different analytical and intervention approaches---a consideration absent from current fairness auditing practice.


\bibliographystyle{splncs04}
\bibliography{main}

\end{document}